\documentclass[journal]{IEEEtran} 
\usepackage{amsmath,amsfonts}
\usepackage{algorithmic}
\usepackage{array}
\usepackage[caption=false,font=normalsize,labelfont=sf,textfont=sf]{subfig}
\usepackage{textcomp}
\usepackage{stfloats}
\usepackage{url}
\usepackage{verbatim}
\usepackage{graphicx}
\def\BibTeX{{\rm B\kern-.05em{\sc i\kern-.025em b}\kern-.08em
    T\kern-.1667em\lower.7ex\hbox{E}\kern-.125emX}}
\usepackage{balance}

\usepackage[table]{xcolor}
\usepackage{graphicx}
\usepackage{multirow}
\usepackage{bbding}
\usepackage{booktabs}
\usepackage{algorithm}
\usepackage{algorithmic}

\newcommand{\eg}{\textit{e}.\textit{g}.}
\newcommand{\etal}{\textit{et al}.}
\newcommand{\ie}{\textit{i}.\textit{e}.}
\newcommand{\etc}{\textit{etc}}

\definecolor{cvprblue}{rgb}{0.21,0.49,0.74}
\usepackage[pagebackref,breaklinks,colorlinks,citecolor=cvprblue]{hyperref}

\begin{document}
\title{Distilling Efficient Vision Transformers from CNNs for Semantic Segmentation}
\author{Xu Zheng, \textit{Student Member, IEEE}, Yunhao Luo, Pengyuan Zhou, Lin Wang$^\dagger$, \textit{Memeber, IEEE}
\thanks{\IEEEcompsocthanksitem X. Zheng is with the AI Thrust, HKUST(GZ), Guangzhou, China. E-mail:  xzheng287@connect.hkust-gz.edu.cn. \\
\IEEEcompsocthanksitem Y. Luo is with Brown University, USA. E-mail: yunhao\_luo@brown.edu. \\
\IEEEcompsocthanksitem P. Zhou is with the University of Science and Technology of China, China. E-mail: pyzhou@ustc.edu.cn. \\
\IEEEcompsocthanksitem L. Wang is with the AI Thrust, HKUST(GZ), Guangzhou, and Dept. of Computer Science and Engineering, HKUST, Hong Kong SAR, China. E-mail: linwang@ust.hk \\
\hfil\break($^\dagger$Corresponding author: Lin Wang)
}
}

\markboth{Journal of \LaTeX\ Class Files,~Vol.~18, No.~9, September~2020}%
{How to Use the IEEEtran \LaTeX \ Templates}

\maketitle

\begin{abstract}
In this paper, we tackle a new problem: \textit { how to transfer knowledge from the pre-trained cumbersome yet well-performed CNN-based model to learn a compact Vision Transformer (ViT)-based model while maintaining its learning capacity}? 
Due to the completely different characteristics of ViT and CNN and the long-existing capacity gap between teacher and student models in Knowledge Distillation (KD), directly transferring the cross-model knowledge is non-trivial.
To this end, we subtly leverage the visual and linguistic-compatible feature character of ViT (\ie, student), and its capacity gap with the CNN (\ie, teacher) and propose a novel CNN-to-ViT KD framework, dubbed \textbf{C2VKD}.
Importantly, as the teacher's features are heterogeneous to those of the student,
we first propose a novel visual-linguistic feature distillation (\textbf{VLFD}) module that explores efficient KD among the aligned visual and linguistic-compatible representations. Moreover, due to the large capacity gap between the teacher and student and the inevitable prediction errors of the teacher, we then propose a pixel-wise decoupled distillation (\textbf{PDD}) module to supervise the student under the combination of labels and teacher's predictions from the decoupled target and non-target classes.
Experiments on \textbf{three} semantic segmentation benchmark datasets consistently show that the increment of mIoU of our method is over \textbf{200\%} of the SoTA KD methods\footnote{Project Page: \url{https://vlislab22.github.io/C2VKD/}}. 
\end{abstract}


\section{Introduction}
\label{sec1}
Although convolutional neural networks (CNNs) have been the primary learning paradigm for image recognition~\cite{he2016deep,chen2021vision}, recent studies have shown that the vision transformer (ViT)\cite{strudel2021segmenter,zheng2021rethinking} has surpassed CNNs in the large-scale data-driven semantic segmentation task, thanks to its unsaturated learning capability and scalability\cite{wolf2020transformers,vaswani2017attention}.
However, a significant challenge with ViT is its high computation and memory costs, particularly when processing high-resolution images for semantic segmentation. Additionally, ViT requires a considerable amount of training data for convergence, leading to higher computational costs than CNNs with similar performance. Therefore, it is crucial to obtain compact ViT models while maintaining their learning capability to reduce computational complexity and improve efficiency.

\begin{figure}[t!]
    \centering
    \includegraphics[width=0.49\textwidth]{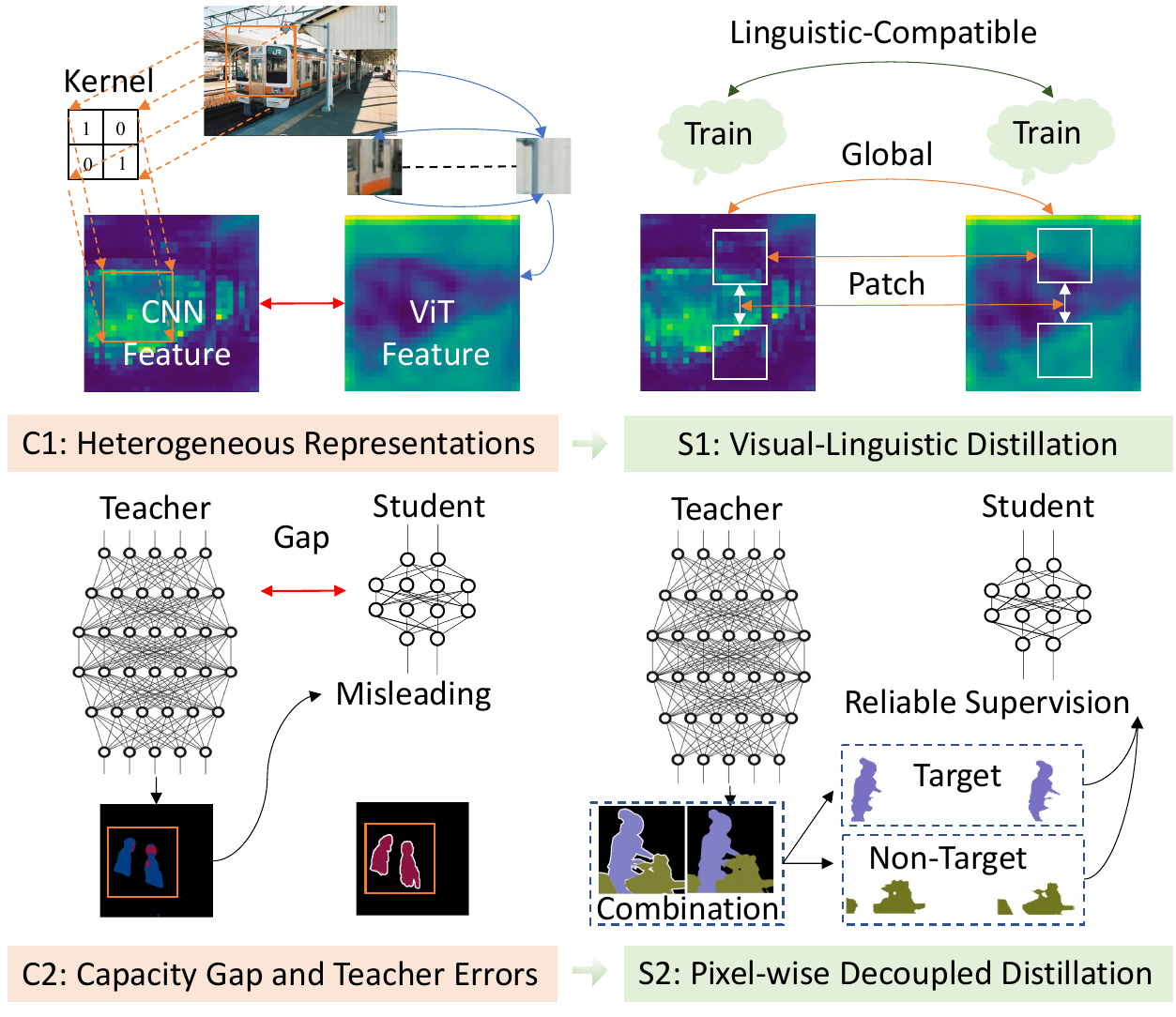}
    \caption{
    Our C2VKD framework effectively addresses two significant challenges: heterogeneous representation between CNN and ViT (\textbf{C1}) and capacity gap and teacher's prediction errors (\textbf{C2}). These challenges are overcome by our proposed solutions: visual-linguistic feature distillation (\textbf{S1}) and pixel-wise decoupled distillation modules (\textbf{S2}).
    }
    \label{coverfig}
\end{figure}

Given the maturity of CNNs in structure design and the existence of numerous pretrained, high-performance CNN models for semantic segmentation~\cite{li2021survey,zheng2022transformer,zhu2023good}, we pose a new problem: \textit{how can we transfer knowledge from these CNN-based models to learn a compact ViT-based model?} This approach enables us to fully leverage the existing CNNs while benefiting from the superiority of the self-attention-based architectures of ViT, thereby improving the efficiency and effectiveness of semantic segmentation tasks.

\begin{figure}[t!]
    \centering
    \includegraphics[width=\linewidth]{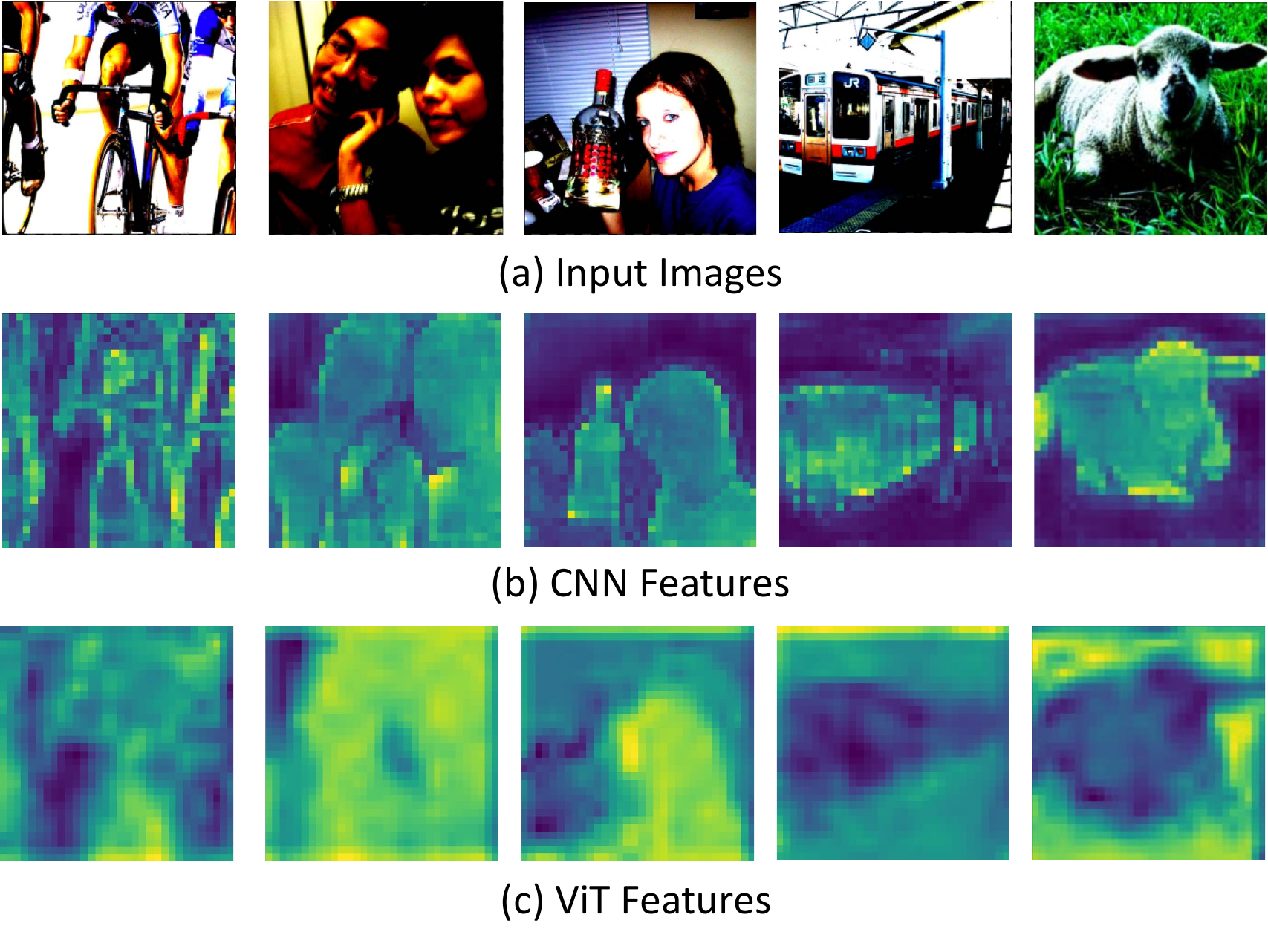}
     \vspace{-22pt}
    \caption{Example visualization of the high-level feature representations from CNN and ViT.}
    \label{feat_compare}
\end{figure}

However, the transfer of knowledge from a CNN-based model (referred to as the teacher) to a ViT-based model (referred to as the student) is a non-trivial task due to the fundamentally different characteristics of ViT. These differences include the receptive fields and lack of prior inductive bias, which have resulted in a long-standing performance gap in knowledge distillation (KD)~\cite{zhou2021convnets,naseer2021intriguing,bai2021transformers, liu2022cross, yang2022vitkd, chi2023prefallkd}. For example, as illustrated in Fig.~\ref{feat_compare}, the high-level features of CNN and ViT exhibit different characteristics, making direct knowledge transfer challenging.
In particular, since the operations in ViT are similar to the transformers used in natural language processing (NLP)\cite{dosovitskiy2020image}, ViT inherits linguistic-compatible characteristics that are brought to the visual domain\cite{radford2021learning}. Neglecting this crucial linguistic-compatible knowledge in the latent space can limit the performance of the student model. Additionally, the visual representations in ViT differ from those in CNNs due to the patch-embedding and self-attention operations, making it difficult to align features. This renders existing KD methods, such as~\cite{liu2019structured} for CNN-based segmentation model compression, less applicable to our problem.

In this paper, we aim to address three key challenges: (\textbf{I}) the heterogeneous receptive fields between the teacher and student features~\cite{gou2021knowledge}, which make it impossible to directly align the representation, as depicted in Fig.~\ref{coverfig}- C1; (\textbf{II}) the significant capacity gap between the CNN-based teacher and ViT-based student~\cite{mirzadeh2020improved,cho2019efficacy}, which can impact the efficacy of knowledge distillation; and (\textbf{III}) the use of separate training for the ViT-based student with the teacher's logits and labels, as done in prior KD methods~\cite{wang2021knowledge,liu2019structured}, which can significantly reduce the student's learning capacity due to prediction errors from the teacher (See Fig.~\ref{coverfig}-C2).

To this end, we propose, to the best of our knowledge, the \textbf{first} and novel KD framework, 
\textbf{C2VKD} (CNN-to-ViT KD), to learn a compact ViT-based student by transferring the feature- and prediction-level knowledge from a CNN-based teacher. 
To tackle the first challenge of heterogeneous receptive fields, we introduce a novel visual-linguistic feature distillation (VLFD) module, as illustrated in Fig.~\ref{coverfig}-S1, which leverages intermediate features to transfer visual and linguistic knowledge simultaneously in the corresponding latent feature spaces (Sec.~\ref{VLFD}). Inspired by CLIP~\cite{radford2021learning} for language-image pre-training, our VLFD module extracts linguistic-compatible characteristics from the teacher's high-level features and aligns them with the student. As the student's inputs are local patches, merely aligning feature maps in a global manner neglects crucial correspondences among patches. Therefore, our VLFD module explores both global-wise and patch-wise visual representations to distill positional and semantic knowledge, which is crucial for semantic segmentation.

To address the other two challenges, we propose a pixel-wise decoupled distillation (PDD) module (See Fig.~\ref{coverfig}-S2) to supervise the student by combining labels and teacher's predictions from the target and non-target classes separately (Sec.~\ref{PDD}). This approach transfers more reliable knowledge and better addresses the problem caused by the model capacity gap~\cite{cho2019efficacy} in knowledge distillation.
By decoupling the distillation process for target and non-target classes, our PDD module reduces the impact of teacher's prediction errors on the student's learning capacity, which is a common challenge in KD. Additionally, our PDD module enables the student to learn more efficiently and effectively from the teacher's predictions, thereby reducing the capacity gap between the teacher and student models.

In summary, our paper makes the following contributions: 
\begin{itemize}
    \item We propose a novel C2VKD framework, which is the first approach for learning a compact ViT-based student by transferring knowledge from a CNN-based teacher.
    \item We present the visual-linguistic feature distillation module, which transfers visual and linguistic feature knowledge simultaneously.
    \item We propose the pixel-wise decoupled distillation module to enable the ViT-based student to learn separately from the target/non-target classes.
    \item Our C2VKD achieves the new state-of-the-art performance on three benchmark datasets for segmentation.
\end{itemize}

\section{Related work}
\subsection{Vision Transformer (ViT)} 
Vision Transformer has been shown to have favorable performance on large-scale data~\cite{han2020survey,yang2021transformer}. However, its performance on limited training data is often unsatisfactory~\cite{chen2021visformer}. To address this issue, many approaches have been proposed to strengthen ViT by introducing well-designed components and schemes~\cite{xu2021vitae,xiao2021early,d2021convit,dai2021coatnet,gehring2017convolutional}. Another approach is to utilize knowledge distillation (KD)\cite{fang2021compressing}. For instance, DeiT\cite{touvron2021training} proposes a knowledge transfer approach to train a ViT-based model in case of insufficient training data.
In addition to considering the visual perspective for ViT~\cite{wu2021visual,liu2021efficient}, some research also explores the vision-language characteristics~\cite{miech2021thinking,kim2021vilt,zhu2021long,shin2022perspectives,li2021align,desai2021virtex}. In particular, Radford \etal~\cite{radford2021learning} demonstrated that obtaining broader supervision from image-paired raw text is a promising auxiliary way for transformer-based vision tasks. Rao \etal~\cite{rao2021denseclip} showed that the ability of vision-language transformation can be applied to dense prediction tasks.
Given the similarities between ViT and transformers in NLP~\cite{dosovitskiy2020image}, ViT is also endowed with linguistic-compatible characteristics. Accordingly, we consider the C2VKD framework from a vision-language perspective.
\subsection{ViT for Semantic Segmentation}
Beyond image recognition, various ViT variants~\cite{ranftl2021vision,wu2021cvt,yuan2021hrformer, zheng2023both, zheng2023look} have been proposed for dense prediction tasks, particularly for semantic segmentation, which requires pyramid features from high-resolution images for better performance. Examples of such ViT variants include Pyramid Vision Transformer~\cite{wang2021pyramid,wang2022pvtv2} and Swin Transformer~\cite{liu2021swin,liu2021swinv2}. Although these variants achieve state-of-the-art performance on various benchmarks, the extra computation required for feature pyramids and self-attention on high-resolution images significantly increases model complexity. To obtain a lightweight network, Xie \etal~\cite{xie2021SegFormer} proposed SegFormer, a relatively simple yet efficient ViT-based network. SegFormer consists of a hierarchically structured transformer encoder, called MiT, and an MLP-based decoder. In this paper, we adopt SegFormer, Pyramid Vision Transformer, and Pyramid Vision Transformer v2 as our ViT-based student networks for the semantic segmentation task. These networks have been shown to achieve state-of-the-art performance on various benchmarks while maintaining relatively low computational complexity.

\begin{figure*}[ht!]
    \centering
    \includegraphics[width=\textwidth]{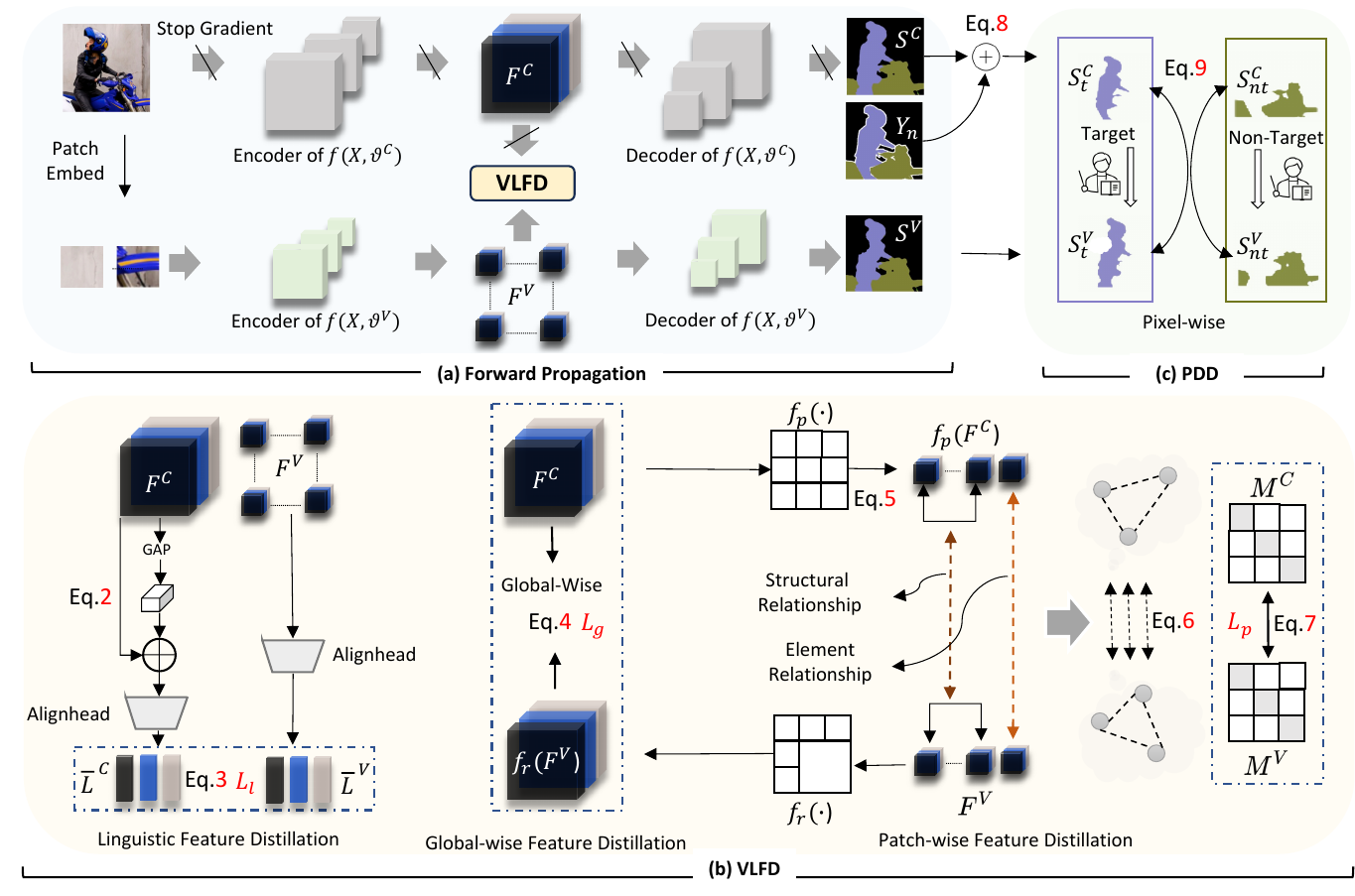}
    \caption{\textbf{Overview of the proposed C2VKD framework}, which consists of a CNN-based teacher $f(X,\theta^{\text{C}})$, a ViT-based student $f(X,\theta^{\text{V}})$ and the KD modules. (a) Forward propagation of the teacher and student networks; (b) The proposed Visual-linguistic Feature Distillation (VLFD) Module; and (c) The proposed Pixel-wise Decoupled Distillation (PDD) module.
    }
\label{fig:overview}
\end{figure*}

\subsection{Knowledge Distillation (KD)} 
Knowledge distillation aims to learn a lightweight student model by transferring the knowledge of a cumbersome yet high-performance teacher model~\cite{hinton2015distilling, chi2023prefallkd}. Mainstream KD methods can be divided into two types~\cite{wang2021knowledge}: KD from the logits~\cite{liu2019knowledge,yang2019snapshot,wen2021preparing,li2017learning} and KD from the intermediate hints~\cite{heo2019comprehensive,tung2019similarity,aguilar2020knowledge,gao2020residual}. The former mostly focuses on optimizing the vanilla KD loss~\cite{zhang2018deep}. To improve the flexibility and efficiency of KD, some works, such as~\cite{zhao2022decoupled}, decouple the target and non-target classes in the output logits. The latter transfers latent representations directly~\cite{adriana2015fitnets} or shares the inner correlation among selected samples~\cite{liu2019structured}. Various KD methods have been proposed for semantic segmentation~\cite{wang2020intra,shu2021channel,michieli2021knowledge,he2019knowledge,michieli2019incremental,holder2021efficient} to obtain compact models that can be used in practical applications, such as autonomous driving~\cite{liu2022transformer,kothandaraman2021domain}. More recently, efforts have been made to compress visual-linguistic CNN models for object detection~\cite{li2020oscar}.

Our work differs from these methods in three aspects: (1) we focus on the challenging task of cross-model KD from a CNN-based teacher to a heterogeneous ViT-based student, which has not been explored before; (2) we propose a VLFD module that aligns the latent representation from visual and linguistic perspectives in C2VKD; and (3) we propose a PDD module that enables the ViT-based student to learn separately from the target/non-target classes in the pixel-level output logits.

\section{Methodology}
An overview of our C2VKD framework is depicted in Fig.~\ref{fig:overview}, which comprises three components: a CNN-based teacher $f(X; \theta^{\text{C}})$, a ViT-based student $f(X; \theta^{\text{V}})$, and a knowledge distillation (KD) module. Our objective is to train a compact $f(X; \theta^{\text{V}})$ that can assign a pixel-wise label $l \in {1,...,K}$ to each pixel $p_{i,j}$ in image $x \in X$ by transferring knowledge from $f(X,\theta^{\text{C}})$. Here, $h$ and $w$ denote the height and width of $x$, and $K$ is the number of classes.
To achieve this goal, we obtain the segmentation confidence maps ($S^{\text{C}}$ and $S^{\text{V}}$) and feature representations ($F^{\text{C}}$ and $F^{\text{V}}$) from the teacher $f(X,\theta^{\text{C}})$ and the student $f(X,\theta^{\text{V}})$, respectively, for a given input image set $X \in \mathbb{R}^{h\times w\times 3 }$, which can be formulated as:
\begin{equation}
    \begin{aligned}
    & (S^{\text{C}}, F^{\text{C}}) = f(X; \theta^{\text{C}}),
    & (S^{\text{V}}, F^{\text{V}}) = f(X; \theta^{\text{V}}),
    \end{aligned}
\end{equation}
as shown in Fig.~\ref{fig:overview} \textcolor{red}{(a)}.
Our key ideas are two folds.
First, because of the identical operations between the student $f(X;\theta^{\text{V}})$ and transformers used in NLP, the intermediate features $F^{\text{V}}$ possess the linguistic-compatible characteristics. Therefore, we propose a visual-linguistic feature distillation module (VLFD) to leverage the intermediate features thoroughly by transferring the visual and linguistic knowledge simultaneously in the corresponding feature spaces.
Specifically, we transform teacher's intermediate features $F^{\text{C}}$ as the linguistic-compatible features to fit the features $F^{\text{V}}$ from the student $f(X;\theta^{\text{V}})$. Meanwhile, as the input unit of $f(X;\theta^{\text{V}})$ is the local patch, merely aligning the features maps in a global manner between the teacher and student may neglect crucial correspondence among patches. 
Therefore,  we explore the visual representation through the global-wise ($G^{\text{C}}$ and $G^{\text{V}}$) and patch-wise($P^{\text{C}}$ and $P^{\text{V}}$) together to distill the inner hint knowledge.
Second, as there exists a considerable capacity gap between the teacher $f(X; \theta^{\text{C}})$ and student $f(X; \theta^{\text{V}})$, and the teacher's predictions, \ie, the segmentation confidence maps $S^{\text{C}}$, may not be precise enough, we propose a pixel-wise decoupled distillation (PDD) module that enables $f(X; \theta^{\text{V}})$ to be supervised by the combination of labels and $P^{\text{C}}$ from target and non-target classes separately.

\subsection{Visual-Linguistic Feature Distillation (VLFD)}
\label{VLFD}
Compared to prior feature KD methods~\cite{wang2021knowledge} that focus on learning a compact CNN-based student from a CNN-based teacher, our approach tackles a more challenging problem: effectively transferring feature representations from a CNN-based teacher $f(X;\theta^{\text{C}})$ to a ViT-based student $f(X;\theta^{\text{V}})$. The basic units of $f(X;\theta^{\text{C}})$ and $f(X;\theta^{\text{V}})$, convolution (Conv) and self-attention (SA), respectively, exhibit opposite behaviors in feature extraction. Conv presents high-pass characteristics, while SA acts like low-pass filters. This makes it impractical to simply align the inner features $F^{\text{C}}$ and $F^{\text{V}}$.

To this end, we propose the VLFD module, which aligns the visual and linguistic-compatible representations simultaneously in the corresponding latent spaces while performing distillation among the aligned features. Our VLFD module, as illustrated in Fig.~\ref{fig:overview} \textcolor{red}{(b)}, comprises three parts: linguistic feature distillation, global-wise feature distillation, and patch-wise feature distillation. We introduce each of these parts in the following sections.

\subsubsection{Linguistic Feature Distillation} 
Given that the image patches in ViT operate on the same principle as the word tokens in NLP, the high-level representations from ViT exhibit linguistic-compatible characteristics, as demonstrated in recent works such as DenseCLIP~\cite{rao2021denseclip}. Therefore, we propose to first align the high-level features from $f(X;\theta^{\text{V}})$ and $f(X;\theta^{\text{C}})$ from a linguistic perspective. To achieve this alignment, we employ an attention pooling component that comprises a global average pooling (GAP) layer and a multi-head self-attention (MHSA) layer. As illustrated in Fig.~\ref{fig:overview} \textcolor{red}{(b)}, the features $F^{\text{C}}$ from the last layer of $f(X; \theta^{\text{C}})$ are first fed into the GAP layer to capture the entire image. Subsequently, the obtained global features are concatenated with $F^{\text{C}}$ to serve as input to the MHSA layer, which further enhances awareness of the entire input:
\begin{equation}
    [\bar{L}^{\text{C}}, L^{\text{C}}] = \text{MHSA}(\textit{Cat}[ \text{GAP}(F^{\text{C}}),F^{\text{C}} ]),
    \label{mhsa}
\end{equation}
where the $\bar{L}^{\text{C}} \in \mathbb{R}^D$ are the global-aware linguistic-compatible features of $f(X,\theta^{\text{C}})$ and finally have inter-relationships with each input element in $F^{\text{C}}$, similar to the \textit{cls} token in transformers for NLP~\cite{rao2021denseclip}.
Also, a component for feature dimension alignment is spliced to the last layer of the backbone of $f(X,\theta^{\text{V}})$ to obtain the $[\bar{L}^{\text{V}}, L^{\text{V}}]$.
Finally, we adopt the KL-Divergence as the linguistic-compatible feature KD loss:
\begin{equation}
\label{ll}
    \mathcal{L}_l = \frac{1}{D} \sum_{d=1}^{D}  \bar{L}^{\text{V}}_{d}\log\frac{\bar{L}^{\text{V}}_{d}}{\bar{L}^{\text{C}}_{d}}.  
\end{equation}

We now elaborate on the details of feature distillation from the visual perspective. Due to the heterogeneity of the input forms (whole image vs. patches) and feature extractors between the CNN-based teacher $f(X;\theta^{\text{C}})$ and ViT-based student $f(X;\theta^{\text{V}})$, we propose global- and patch-wise feature distillation to align the visual representations.

\subsubsection{Global-wise feature distillation}
Although the high-dimensional features obtained by $f(X;\theta^{\text{C}})$ and $f(X;\theta^{\text{V}})$ in Fig.\ref{fig:overview} \textcolor{red}{(b)} are explicitly different~\cite{raghu2021vision}, they share implicit commonalities due to the same input. In the deeper layers of $f(X;\theta^{\text{C}})$, the receptive fields of the representations grow larger and cover a significant portion of the image.
Intuitively, we propose to transfer the knowledge in the high-level features from the last layer of $f(X;\theta^{\text{C}})$ to $f(X;\theta^{\text{V}})$.
Moreover, as the $F^{\text{V}}$ is obtained based on the patches of an input image, we design a reverse function $f_r(\cdot)$ to rebuild $F_{\text{ViT}}$ to a feature map.
As such, we can measure the global-wise feature discrepancy $\mathcal{L}_g$ between $F^{\text{V}}$ and $F^{\text{C}}$ by computing the KL-Divergence between $f_r$($F^{\text{V}}$) and $F^{\text{C}}$, which can be formulated as:
\begin{equation}
\label{lg}
        \mathcal{L}_g = f_r(F^{\text{V}})\log \frac{f_r(F^{\text{V}})}{F^{\text{C}}}.
\end{equation}

Because the global-wise characteristics are only a small representation partition of the whole sequence, more critical positions, and semantic inter-relationships exist across elements that are mapped from the individual patches in the same input image.
Therefore, we also exploit the high-level representations $F^{\text{V}}$ and $F^{\text{C}}$ from a patch-wise perspective. We now describe details as follows.

\subsubsection{Patch-wise feature distillation}
Theoretically, due to the input being a sequence of image patches, the internal features extracted naturally pose sequential properties. 
The process of $f(X;\theta^{\text{V}})$ mapping the split input $f_p(x)$ to a sequence of continuous representations in Fig.~\ref{fig:overview} \textcolor{red}{(b)} is:
\begin{equation}
\label{patch}
    (n_1,n_2,n_3,n_4,...,n_T) = f(f_p(x);\theta^{\text{V}}),
\end{equation}
where $n_t \in \mathbb{R}^{Z}$ denotes the high-level representation vector of $t^{th}$ patch of input $x$, $f_p$($\cdot$) is the patch partition operation. In contrast to the global-wise feature KD which takes $\begin{vmatrix}
  n_1,n_2 \\
  n_3,n_4
\end{vmatrix}$
(\ie, $T$ = 4) as a whole distribution, we further explore the pair-wise positional and semantic inter-relationships among $(n_1,n_2,n_3,n_4)$, which are important for semantic segmentation. Specifically, we split the high-level features of the teacher by the same partition strategy on $f(X;\theta^{\text{V}})$'s input to get $f_p(F^{\text{C}})$. 
Therefore, the idea of patch-wise feature KD can be formulated as:
\begin{equation}
\label{affinity}
   \begin{vmatrix}
  n_1& \longleftrightarrow  &n_2 \\
  \updownarrow & \times    & \updownarrow\\
  n_3& \longleftrightarrow  &n_4
\end{vmatrix}\Leftrightarrow \begin{vmatrix}
  n_1^{'}& \longleftrightarrow  &n_2^{'} \\
  \updownarrow & \times    & \updownarrow\\
  n_3^{'}& \longleftrightarrow  &n_4^{'}
    \end{vmatrix} ,
\end{equation}
where the $(n_1,n_2,n_3,n_4)$ and $(n_1^{'},n_2^{'},n_3^{'},n_4^{'})$ are instances of the $f_p(F^{\text{V}})$ and the $f_p(F^{\text{C}})$ with $T = 4$, the arrows and crosses denote the relationships among $n_t$. Both of the individual elements $n_t$ and the relationships of $f_p(F^{\text{V}})$ are imposed to fit with $f_p(F^{\text{C}})$'s for the structure-wise and element-wise KD.
Intuitively, these elements and their internal relationships are described as nodes and edges in a mapped graph. As such, the affinity matrix $M$ can be easily obtained by matrix multiplication of $(n_1,n_2,n_3,n_4)$ and $(n_1,n_2,n_3,n_4)^{tr}$.

In Eq.~\ref{affinity}, $M^{\text{C}}$ and $M^{\text{V}}$ are the patch-wise affinity matrices for $f(X;\theta^{\text{C}})$ and $f(X;\theta^{\text{V}})$, and the diagonal and off-diagonal elements refer to the correspondence affinity within the patches $n_t$ and the relation among $(n_1,n_2,...,n_T)$, respectively. Finally, the patch-wise feature KD loss is formulated by computing the Mean Square Error (MSE) between the affinity matrices:
\begin{equation}
     \mathcal{L}_p = \frac{1}{T^2 \times Z} \sum_{t=1}^{T}(M^{\text{C}}_t - M^{\text{V}}_t)^2
\end{equation}

\begin{figure*}[t!]
    \centering
    \includegraphics[width=\textwidth]{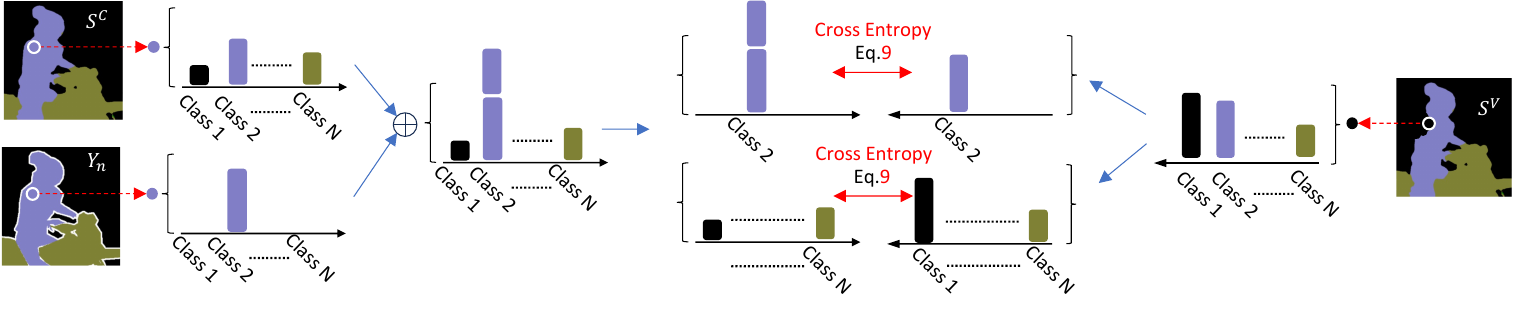}
    \vspace{-24pt}
    \caption{Illustration of the proposed Pixel-wise Decoupled Distillation (PDD) module.}
\label{fig:pdd}
\end{figure*}

\subsection{Pixel-wise Decoupled Distillation (PDD)}
\label{PDD}
In prior KD methods, such as~\cite{hinton2015distilling}, the student network is trained jointly using labels and the teacher's output logits. However, using these methods naively can degrade KD efficiency if the teacher network $f(X,\theta^{\text{C}})$ makes mistakes, particularly for dense prediction tasks like semantic segmentation. 
To address this problem, we propose the Pixel-wise Decoupled Distillation (PDD) module, which serves two distinct roles.
Firstly, PDD combines the label $Y$ and output logits $S^{\text{C}}$ to provide soft supervision for $f(X;\theta^{\text{V}})$. Secondly, it decouples the target class and non-target class distributions for the ViT-based student $f(X;\theta^{\text{V}})$ using pixel-wise decoupled distillation, inspired by \cite{zhao2022decoupled} for classification.

\begin{algorithm}[ht!]
	\caption{The proposed C2VKD framework} 
	\label{alg} 
	\begin{algorithmic}[1]
	    \STATE \textbf{Input}: $x$, maximum iteration: $T$, teacher: $f(X;\theta^C)$, student: $f(X;\theta^V)$; \vspace{4pt}
	    \STATE  \textbf{Initialization}: Set $\theta^C$ with per-trained ResNet-101 and $\theta^V$ with pre-trained SegFormer; \vspace{4pt}
	    \FOR{t $\xleftarrow[]{}$ 1 to $T$ do} \vspace{4pt} 
    	    \STATE  $(S^C, F^V) = f(x; \theta^C),(S^V, F^V) = f(x; \theta^V)$; \vspace{4pt}
    	    \STATE  $\bar{L}^C$ = Attention-Pooling($F^C$), $\bar{L}^V$ = Align($F^V$); \vspace{4pt}
    	    \STATE  $\mathcal{L}_l$ = KL($\bar{L}^V,\bar{L}^C$); \vspace{4pt}
    	    \STATE  $\mathcal{L}_g$ = KL($f_r(F^V)$,$F^C$); \vspace{4pt}
    	    \STATE  $M^C$ = Patch-Affinity$(f_p(F^C))$, \vspace{4pt} \\ $M^V$ = Patch-Affinity$(f_p(F^V))$; \vspace{4pt}
    	    \STATE  $\mathcal{L}_p$ = MSE($M^C,M^V$); \vspace{4pt}
    	    \STATE  (($S^C_t$, $S_{nt}^C$), ($S_t^V$, $S_{nt}^V$)) = Decoupled($S^C$,$S^V$); \vspace{4pt}
    	    \STATE  $\mathcal{L}_d$ = KL($S_t^C$,$S_t^V$) + KL($S_{nt}^C$,$S_{nt}^V$); \vspace{4pt}
    	    \STATE  $\mathcal{L}_{all}$ = $\mathcal{L}_d + \lambda_g * \mathcal{L}_g + \lambda_p * \mathcal{L}_p + \lambda_l * \mathcal{L}_l$; \vspace{4pt}
    	    \STATE  Back propagation for $\mathcal{L}_{all}$; \vspace{4pt}
    	    \STATE  Update parameter set $\theta^V$; \vspace{4pt}
    	 \ENDFOR \vspace{4pt}
	    \STATE  \textbf{return}  $\theta^V$ \vspace{4pt}
	    \STATE  \textbf{End}.
	\end{algorithmic} 
\end{algorithm}

Specifically, as illustrated in Fig.~\ref{fig:pdd}, for a pixel $p_{i,j}$ in the segmentation confidence maps $S$, the predicted logits can be denoted as $p_{i,j} = [l_1, l_2, ..., ... l_i, ..., l_K]$, where $l_k$ is the probability of the $k$-th class. We separate the pixel-wise predictions relevant and irrelevant to the target class ($cls$-th) and non-target classes into binary probabilities as follows:
\begin{equation}
\label{decouple}
    l_t = l_K, \; \; \;l_{nt} =  {\textstyle \sum_{k=1}^{cls-1}} l_k + {\textstyle \sum_{k=cls+1}^{N}} l_k, \; \; \; [l_t, l_{nt}] \in \mathbb{R}^{2}.
\end{equation}

Then we use segmentation maps $S^t$ and $S^{nt}$ to show target and non-target class binary probability maps. As such, We use KL-Divergence to achieve PDD:
\begin{equation}
\label{ld}
\mathcal{L}_d =  \frac{1}{|P|}( \alpha \cdot S^{\text{V}}_t log(\frac{S^{\text{V}}_t}{S^{\text{C}}_t + Y_n^{t}}) + \beta \cdot S^{\text{V}}_{nt}log(\frac{S^{\text{V}}_{nt}}{S^{\text{C}}_{nt} + Y_n^{nt}})),
\end{equation}
where $Y_n$ denotes the corresponding ground-truth label for the input image $x$, $\alpha$ and $\beta$ are trade-off weights.

Overall, the proposed framework is shown in Algorithm~\ref{alg}, the total training objective contains four losses: linguistic feature distillation loss ($L_l$), global-wise feature distillation loss ($L_g$), patch-wise feature distillation loss ($L_p$), and pixel-wise decoupled distillation loss ($L_d$), and the total loss is as follows:
\begin{equation}
\label{lall}
    \mathcal{L}= \mathcal{L}_d + \lambda_g \cdot \mathcal{L}_g + \lambda_p \cdot \mathcal{L}_p + \lambda_l \cdot \mathcal{L}_l,
\end{equation}
where the $\lambda_g$, $\lambda_p$, and $\lambda_l$ are the trade-off weight to balance the four different losses.

\begin{table*}[t!]
\renewcommand{\tabcolsep}{12pt}
\vspace{-8pt}
\footnotesize
\centering
\resizebox{\textwidth}{!}{
\begin{tabular}{l|c|c|c|c|c|c}\midrule
Method &Backbone & Seg-Head & \#Para(M) & FLOPs(G) & mIoU(\%) &$\triangle$ \\\midrule
- &T: ResNet-101 & DeepLabV3+   &58.8M       &79.16G    &75.77 &-         \\\midrule
- &   &    &  &  &69.51&base \\
KD~\cite{hinton2015distilling} &   &    &   &  &70.05&+0.54 $\uparrow$ \\
IFVD~\cite{wang2020intra} & MiT-B0~\cite{xie2021SegFormer}  & SegFormer~\cite{xie2021SegFormer}   &3.8M   &6.96G   &69.77&+0.26$\uparrow$ \\
CD~\cite{shu2021channel} &   &   &  &   &70.07&+0.56$\uparrow$ \\
Ours &   &   &  & &\textbf{70.76} & \textbf{+1.25$\uparrow$}\\ \midrule
- &   &   &   &   &75.48&base \\
KD~\cite{hinton2015distilling} &   &  &   &   &75.77&+0.29$\uparrow$ \\
IFVD~\cite{wang2020intra} & MiT-B1~\cite{xie2021SegFormer}  & SegFormer~\cite{xie2021SegFormer}   &16.03M   &27.05G   &75.67&+0.19$\uparrow$ \\
CD~\cite{shu2021channel} &   &  &  &   &75.82&+0.34$\uparrow$ \\
Ours &  &   &   &  &\textbf{76.33} &\textbf{+0.85$\uparrow$} \\\midrule
- &   &    &   & &63.52 &base \\
KD~\cite{hinton2015distilling} &   &    &   &   & 64.88 & +1.36$\uparrow$ \\
IFVD~\cite{wang2020intra} &  PVT-T~\cite{wang2022pvt}  &  FPN~\cite{kirillov2019panoptic}   & 13.73M   & 12.54G   & 64.91 & +1.39$\uparrow$ \\
CD~\cite{shu2021channel} &  &    &  &   & 63.38 &-0.14$\downarrow$ \\
Ours &   &    &   &   & \textbf{65.53}& \textbf{+2.01}\\
\midrule
- &   &    &   & &65.77 &base \\
KD~\cite{hinton2015distilling} &   &    &   &   & 66.28  & +0.51$\uparrow$ \\
IFVD~\cite{wang2020intra} & PVTv2-B0~\cite{wang2022pvtv2}  &  FPN~\cite{kirillov2019panoptic}   & 3.46M   & 3.42G   & 64.80  &-1.27$\downarrow$ \\
CD~\cite{shu2021channel} &  &    &  & & 66.07 & +0.30$\uparrow$ \\
Ours &   &    &   &   & \textbf{66.98}& \textbf{+1.21}\\
\bottomrule
\end{tabular}}
\vspace{4pt}
\caption{Comparison with SoTA KD methods on the \textit{PASCAL VOC 2012} val set under different backbones and segmentation heads.}
\label{voc}
\end{table*}

\section{Experiments and Evaluation}
\label{experiments}
\subsection{Datasets}
\noindent \textit{PASCAL VOC 2012} dataset is a fine-annotated dataset that contains 21 classes. The standard training set and validation set comprise 1464 and 1449 object-centered images, respectively.

\noindent \textit{Cityscapes} is a dataset that contains urban street scenes from 50 different cities with high-quality pixel-wise annotations. The official split consists of 5000 finely annotated images, of which 2975/500/1525 are used for \textit{train/val/test}.

\noindent \textit{ADE20K} covers 150 fine-grained semantic concepts and comprises 20210 training images and 2000 validation images.

\subsection{Evaluation}
We take the mean Intersection-over-Union (mIoU) as the evaluation metric for semantic segmentation. Also, we report the network parameters and the sum of floating point operations (FLOPs) on a fixed input size to show the model size and complexity.
The student networks are evaluated on \textit{PASCAL VOC 2012}, \textit{Cityscapes} and \textit{ADE20K} validation sets (1449/500/2000 images). For \textit{PASCAL VOC 2012} and \textit{ADE20K}, we resize and center-crop validation images to 512 $\times$ 512; for \textit{Cityscapes}, we use sliding-window test by cropping 512 $\times$ 512 windows during inference.

\subsection{Implementation details}
The proposed C2VKD framework is built using Pytorch and trained on 4 $\times$ NVIDIA GPUs. 
We initialize the encoder with ImageNet-1K pretrained weight and randomly initialize the decoder (segmentation head). During training, we apply random horizontal flipping and random cropping to 512 $\times$ 512 for three datasets.
We only add extra components (attention pooling$/$align head) to the teacher and student networks when training and no additional operations are added during inference. For all the experiments, we choose the typical segmentation head DeepLabv3+~\cite{chen2018encoder_deeplabv3p} with ResNet-101 as the cumbersome yet high-performance CNN-based teacher. We adopt the SegFormer~\cite{xie2021SegFormer}, PVT~\cite{wang2022pvt} and PVTv2~\cite{wang2022pvtv2} as the ViT-based student. 
We use a batch size of 24 and we train the student models using AdamW optimizer for 40K, 50K for \textit{PASCAL VOC 2012} and \textit{Cityscapes} respectively. As in \cite{xie2021SegFormer}, we set the initialized learning rate as 0.00006 and use a \textit{poly} learning rate schedule with power factor 1.0. \textit{\textbf{For a fair comparison in knowledge distillation efficiency, we do not use any widely-used tricks, \eg, auxiliary segmentation head loss, for all the teacher/student networks in this paper. All the reported numbers of all the comparison KD methods are obtained with the original official open sourced codes\footnote{https://github.com/YukangWang/IFVD} \footnote{https://github.com/irfanICMLL/TorchDistiller/tree/main/SemSeg-distill }.}}

\subsection{Experimental Results}
\label{experiment results}
Our C2VKD consistently outperforms the prior KD methods~\cite{hinton2015distilling, wang2020intra, shu2021channel} with different ViT variants, including SegFormer~\cite{xie2021SegFormer}, PVT~\cite{wang2022pvt} and PVTv2~\cite{wang2022pvtv2}, on all the three semantic segmentation benchmarks.

\begin{figure*}[ht!]
    \centering
    \includegraphics[width=\textwidth]{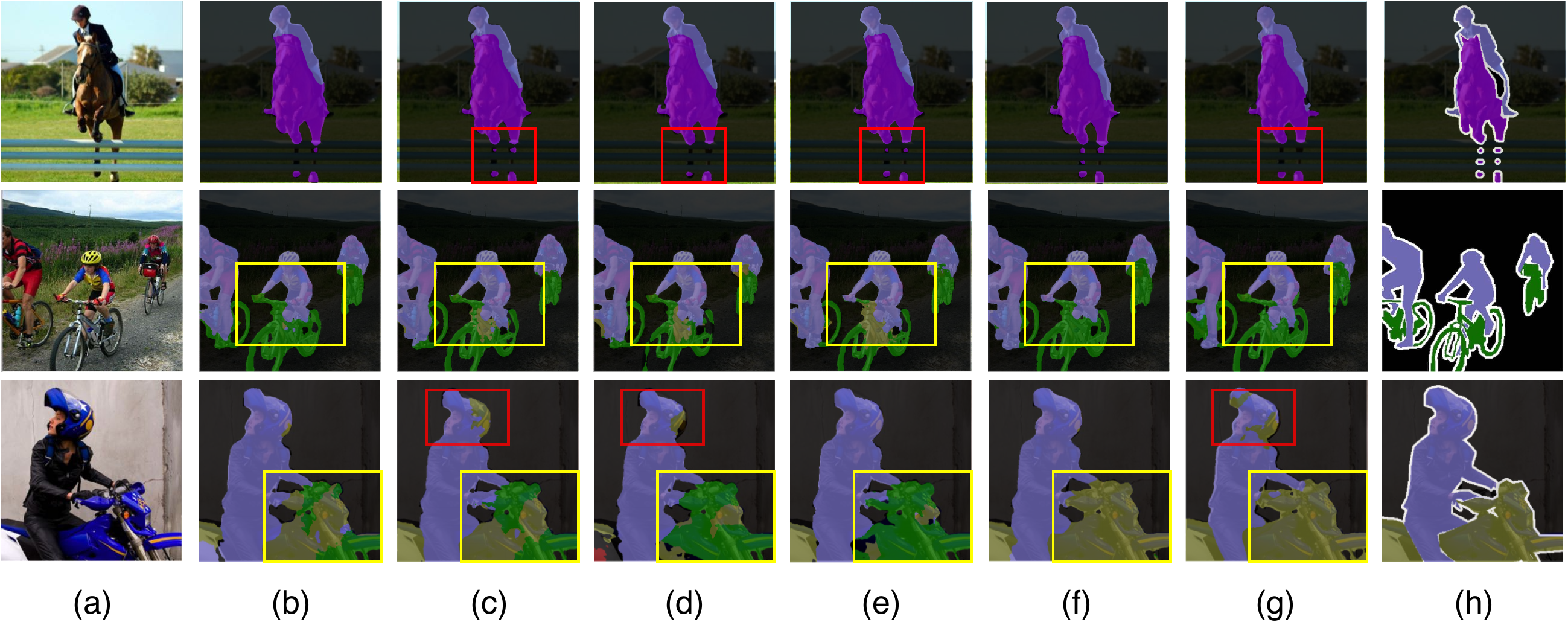}
    \caption{\textbf{Example results from PASCAL VOC 2012}. (a) input, (b) fully supervised by CE loss, (c) CD~\cite{shu2021channel}, (d) KD, (e) IFVD~\cite{wang2020intra}, (f) ours, (g) teacher's (ResNet101), and (h) ground truth.}
    \label{vocresults}
\end{figure*}

\begin{table}[t!]
\renewcommand{\tabcolsep}{2pt}
\vspace{-8pt}
\footnotesize
\centering
\resizebox{\linewidth}{!}{
\begin{tabular}{l|c|c|c|c}\midrule
Method &Backbone & Seg-Head & mIoU(\%) &$\triangle$ \\\midrule
- &T: ResNet-101 & DeepLabV3+    &77.48 &-         \\\midrule
- &   &   &72.52 & base \\
KD~\cite{hinton2015distilling} &   &  &72.66 &+0.14 $\uparrow$ \\
IFVD~\cite{wang2020intra} & MiT-B0~\cite{xie2021SegFormer}  & SegFormer~\cite{xie2021SegFormer}    &72.09&-0.43$\uparrow$ \\
CD~\cite{shu2021channel} &   &   &72.58&+0.06$\uparrow$ \\
Ours &   & &\textbf{73.58} & \textbf{+1.06$\uparrow$}\\ \midrule
- &   &   &73.34&base \\
KD~\cite{hinton2015distilling} &   &   &73.36&+0.02$\uparrow$ \\
IFVD~\cite{wang2020intra} & MiT-B1~\cite{xie2021SegFormer}  & SegFormer~\cite{xie2021SegFormer}    &73.63&+0.29$\uparrow$ \\
CD~\cite{shu2021channel} &   & &72.20&-1.43$\uparrow$ \\
Ours & &  &\textbf{75.25} &\textbf{+1.91$\uparrow$} \\\midrule
- &  & &64.93 &base \\
KD~\cite{hinton2015distilling} &   &   & 65.27 & +0.34$\uparrow$ \\
IFVD~\cite{wang2020intra} &  PVT-T~\cite{wang2022pvt}  &  FPN~\cite{kirillov2019panoptic}   & 63.38 & -0.14$\uparrow$ \\
CD~\cite{shu2021channel} &  &  & 65.09 &+0.16$\downarrow$ \\
Ours &   &    & \textbf{66.85}& \textbf{+1.92}\\
\midrule
- &    & &65.25 &base \\
KD~\cite{hinton2015distilling} &   &   & 65.53  & +0.28$\uparrow$ \\
IFVD~\cite{wang2020intra} & PVTv2-B0~\cite{wang2022pvtv2}  &  FPN~\cite{kirillov2019panoptic} & 65.93  &+0.68$\downarrow$ \\
CD~\cite{shu2021channel} &   & & 66.07 & +0.82$\uparrow$ \\
Ours &   &    & \textbf{67.56}& \textbf{+2.31}\\
\bottomrule
\end{tabular}}
\vspace{4pt}
\caption{Comparison with SoTA KD methods on the \textit{Cityscapes} val set under different backbones and segmentation heads.}
\label{city}
\end{table}

\subsubsection{Results on PASCAL VOC 2012}

Table \ref{voc} reports the quantitative results of four backbone models and three segmentation heads on PASCAL VOC 2012. Our  C2VKD framework outperforms the fully supervised student without any knowledge distillation, achieving a significant improvement in semantic segmentation performance by \textbf{+1.25\%}, \textbf{+0.85\%}, \textbf{+2.01\%}, and \textbf{+1.21\%} in mIoU with SegFormer-B0, SegFormer-B1, PVT-T, and PVTv2-B0, respectively. In contrast, the state-of-the-art feature-based KD methods, IFVD~\cite{wang2020intra} and CD~\cite{shu2021channel}, which outperform KD~\cite{hinton2015distilling} in distilling the CNN-based student from the CNN-based teacher, are severely constrained in learning an efficient ViT-based student due to the first and third challenges mentioned in Sec.~\ref{sec1}. Our proposed C2VKD method better addresses these challenges with our proposed knowledge distillation modules, resulting in a significant performance enhancement. This indicates that our C2VKD framework tackles the second challenge more effectively by reducing the capacity gap from the ViT teacher to the CNN student.

Fig.\ref{vocresults} presents the visual outcomes of the PASCAL VOC dataset. The red boxes denote the longstanding issue in KD, where the student underperforms despite the prediction errors from the teacher, as observed in CD\cite{shu2021channel}(c) and IFVD~\cite{wang2020intra}(d). This issue arises due to the difficulty in transferring knowledge from a large, complex teacher network to a smaller student network. Additionally, the yellow boxes highlight the incorrect segmentation region of the student solely trained by the CE loss. All the compared KD methods inherit and even exacerbate the wrong predictions (red regions), which can lead to a sub-optimal segmentation performance. In contrast, our C2VKD framework effectively mitigates this issue by leveraging our proposed knowledge distillation modules. This leads to a significant improvement in performance, as illustrated in box (f), where our method achieves the best results among all the compared methods. Our proposed method achieves this by addressing the capacity gap between the teacher and student networks and reducing the negative impact of incorrect predictions. Overall, our C2VKD framework demonstrates its effectiveness in overcoming the limitations of existing KD methods and improving the segmentation performance of the student network.

\begin{table}[t!]
\renewcommand{\tabcolsep}{2pt}
\vspace{-8pt}
\footnotesize
\centering
\resizebox{\linewidth}{!}{
\begin{tabular}{l|c|c|c|c}\midrule
Method &Backbone & Seg-Head & mIoU(\%) &$\triangle$ \\\midrule
- &T: ResNet-101 & DeepLabV3+    &40.16 &-         \\\midrule
- &   &   &35.86&base \\
KD~\cite{hinton2015distilling} &   &  &36.37&+0.51 $\uparrow$ \\
IFVD~\cite{wang2020intra} & MiT-B0~\cite{xie2021SegFormer}  & SegFormer~\cite{xie2021SegFormer}    &36.45&+0.59$\uparrow$ \\
CD~\cite{shu2021channel} &   &   &37.01&+1.15$\uparrow$ \\
Ours &   & &\textbf{37.48} & \textbf{+1.62$\uparrow$}\\ \midrule
- &   &   &37.04&base \\
KD~\cite{hinton2015distilling} &   &   &38.36&+1.32$\uparrow$ \\
IFVD~\cite{wang2020intra} & MiT-B1~\cite{xie2021SegFormer}  & SegFormer~\cite{xie2021SegFormer}    &39.52&+2.48$\uparrow$ \\
CD~\cite{shu2021channel} &   & &39.08&+2.04$\uparrow$ \\
Ours & &  &\textbf{40.91} &\textbf{+3.87$\uparrow$} \\\midrule
- &  & &33.49 &base \\
KD~\cite{hinton2015distilling} &   &   & 34.11 & +0.62$\uparrow$ \\
IFVD~\cite{wang2020intra} &  PVT-T~\cite{wang2022pvt}  &  FPN~\cite{kirillov2019panoptic}   & 34.77 & +1.28$\uparrow$ \\
CD~\cite{shu2021channel} &  &  & 34.02 &+0.53$\downarrow$ \\
Ours &   &    & \textbf{35.44}& \textbf{+1.95}\\
\midrule
- &    & &33.98 &base \\
KD~\cite{hinton2015distilling} &   &   & 33.54  & -0.44$\uparrow$ \\
IFVD~\cite{wang2020intra} & PVTv2-B0~\cite{wang2022pvtv2}  &  FPN~\cite{kirillov2019panoptic} & 34.98  &+1.00$\downarrow$ \\
CD~\cite{shu2021channel} &   & & 34.23 & +0.25$\uparrow$ \\
Ours &   &    & \textbf{36.02}& \textbf{+2.04}\\
\bottomrule
\end{tabular}}
\vspace{4pt}
\caption{Comparison with SoTA KD methods on the \textit{ADE20K} val set under different backbones and segmentation heads.}
\label{ade}
\end{table}
\begin{figure*}[t!]
    \centering
    \includegraphics[width=\textwidth]{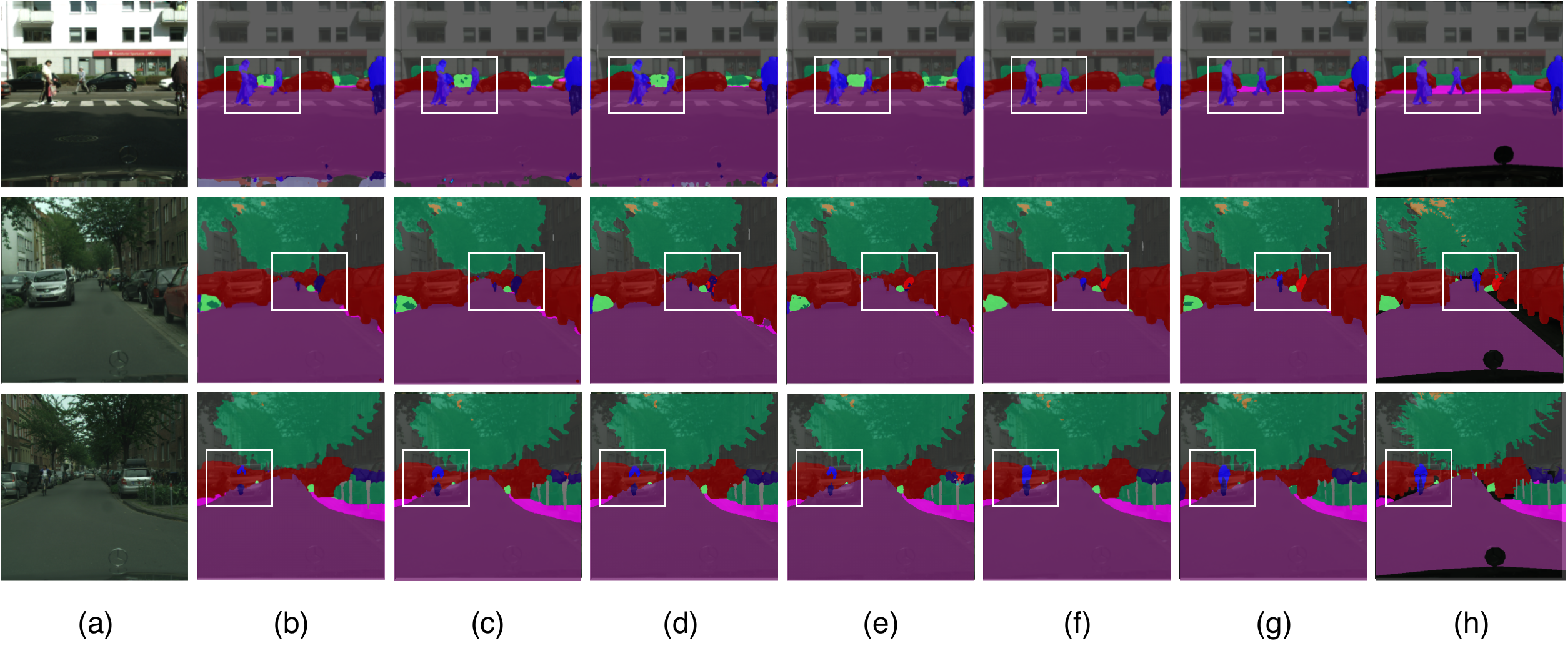}
    \caption{\textbf{Example results from Cityscapes}. (a) input, (b) fully supervised by CE loss, (c) CD~\cite{shu2021channel}, (d) KD, (e) IFVD~\cite{wang2020intra}, (f) ours, (g) teacher's (ResNet101), and (h) ground truth.}
    \label{cityresults}
\end{figure*}

\subsubsection{Results on Cityscapes}

Tab.~\ref{city} reports the quantitative outcomes of the Cityscapes dataset. Our proposed C2VKD consistently outperforms the existing SoTA methods with all the backbone models. Notably, with the efficient PVTv2-B0 model, our C2VKD achieves the largest performance gain of \textbf{+2.31\%} mIoU compared to the baseline.

The Cityscapes dataset comprises street scene images that are more complex than those with fewer classes/objects in the PASCAL VOC dataset. As a result, critical details such as pedestrians and street corners are often overlooked in semantic segmentation, making it challenging for prior KD methods to achieve high efficiency. As shown in Fig.~\ref{cityresults}, the student without KD (b) and compared KD methods (c,d,e) fail to correctly segment the driving scene details (white boxes). However, our proposed C2VKD overcomes these challenges and exhibits the best performance in segmenting these small objects (f), which are crucial in practical applications. These results indicate that our C2VKD method is the most robust in transferring knowledge from the CNN-based teacher to the heterogeneous ViT-based student.

\begin{figure*}[t!]
    \centering
    \includegraphics[width=\textwidth]{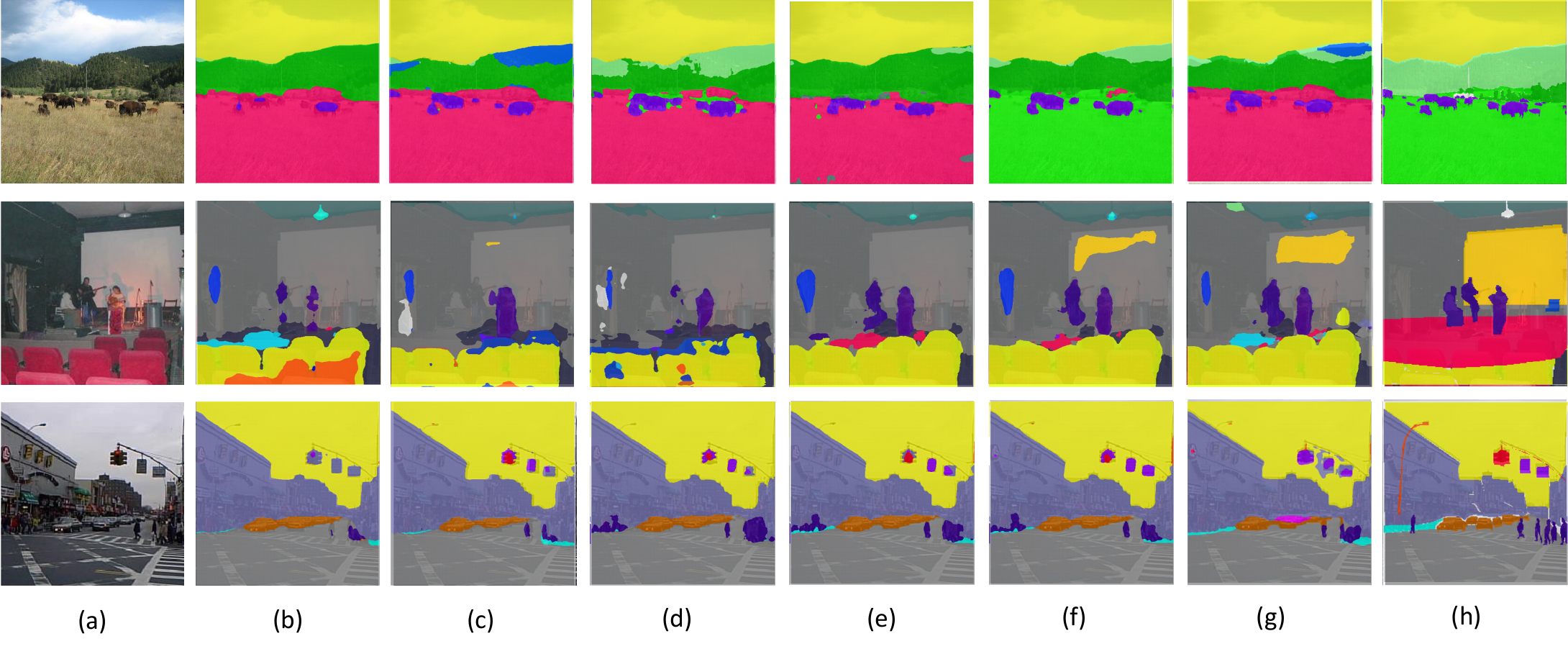}
    \caption{\textbf{Example results from ADE20K dataset}. (a) input, (b) fully supervised by CE loss, (c) CD~\cite{shu2021channel}, (d) KD, (e) IFVD~\cite{wang2020intra}, (f) ours, (g) teacher's (ResNet101), and (h) ground truth.}
    \label{aderesults}
\end{figure*}

\subsubsection{Results on ADE20K}

Tab.~\ref{ade} presents the quantitative outcomes of the ADE20K dataset. Despite being a much larger dataset than PASCAL VOC and Cityscapes, our proposed C2VKD method outperforms the existing state-of-the-art methods with all the backbone models. Notably, with the powerful SegFormer-B1 model, our C2VKD achieves the largest performance gain of \textbf{+3.87\%} mIoU compared to the baseline.

As shown in Fig.~\ref{aderesults}, the student without any KD methods (b), as well as those with other KD methods (c, d, and e), fail to correctly segment the grass (first row of Fig.~\ref{aderesults}). In contrast, our proposed C2VKD method achieves the best performance, demonstrating its robustness in larger datasets such as ADE20K.

\section{Ablation study and Analysis}

 \begin{figure*}[!t]
    \centering
    \includegraphics[width=\textwidth]{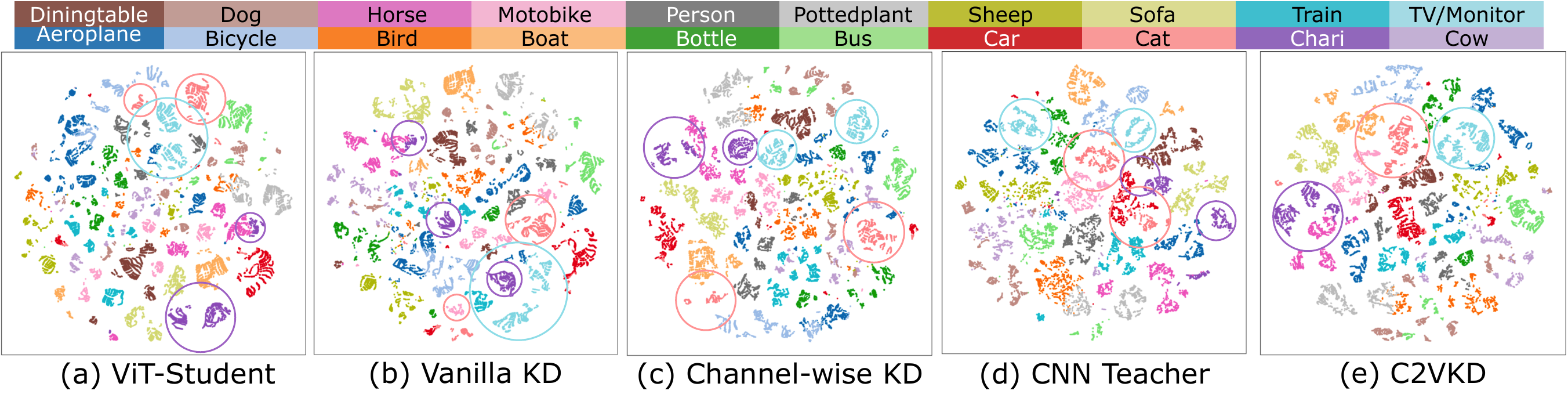}
    \caption{\textbf{TSNE visualization teacher and students learned under different KD methods}. We outline some classes with circles in their colors for a clearer view. In (e), our method encourages better intra-class compactness and inter-class sparseness.
    }
    \label{tsne}
\end{figure*}

\subsection{The effectiveness of VLFD module}
As illustrated in Fig.~\ref{feat_compare}, the features extracted from CNN and ViT exhibit obvious differences in terms of receptive fields and range dependencies. Moreover, Fig.~\ref{tsne} presents the qualitative results of the extracted heterogeneous features. The channel-wise KD method, which aligns cross-model representations in a channel-wise manner, fails to distinguish the discriminative features for each category, resulting in worse segmentation performance for the ViT student. These findings validate that directly applying existing methods to our CNN-to-ViT KD problem is not feasible, as mentioned in Fig.~\ref{coverfig}.

As illustrated in Fig.~\ref{tsne}, the channel-wise KD method fails to distinguish the discriminative features for each category, leading to worse segmentation results for the ViT student. This finding validates that directly applying existing methods to our CNN-to-ViT KD problem is not feasible. In contrast, our proposed VLFD method aligns cross-model features through linguistic and visual distillation, leveraging the specific characteristics of both models.

Tab.~\ref{abloss} presents the effectiveness of different feature distillation methods on the PASCAL VOC and Cityscapes datasets. Our linguistic feature distillation method plays a more significant role in the PASCAL VOC dataset, which has fewer classes and typically contains only one main/center object in an image. This indicates that capturing linguistic-compatible features is easier on this dataset. Specifically, our linguistic feature distillation method brings a mIoU gain of +0.33$\%$, which is higher than the global-wise (+0.15$\%$) and patch-wise feature distillation (+0.22$\%$) under the same settings. The highest mIoU gain validates the superiority of our proposed linguistic feature distillation method. In contrast, Cityscapes contains objects in multiple scales and more than five classes co-exist in an image, making it more challenging to extract linguistic-compatible features (+0.23$\%$ mIoU increment). Overall, our proposed linguistic feature distillation consistently contributes positively to KD in the feature space.

As presented in Tab.~\ref{abloss}, our proposed patch-wise feature distillation consistently achieves mIoU gains on the PASCAL VOC and Cityscapes datasets (by +0.15$\%$ and +0.41$\%$, respectively) compared to the global-wise feature distillation. This result validates the rationality of our patch-wise feature distillation method, which is subtly designed by considering the specific characteristics of ViT.

\begin{figure*}[t!]
    \centering
    \includegraphics[width=\textwidth]{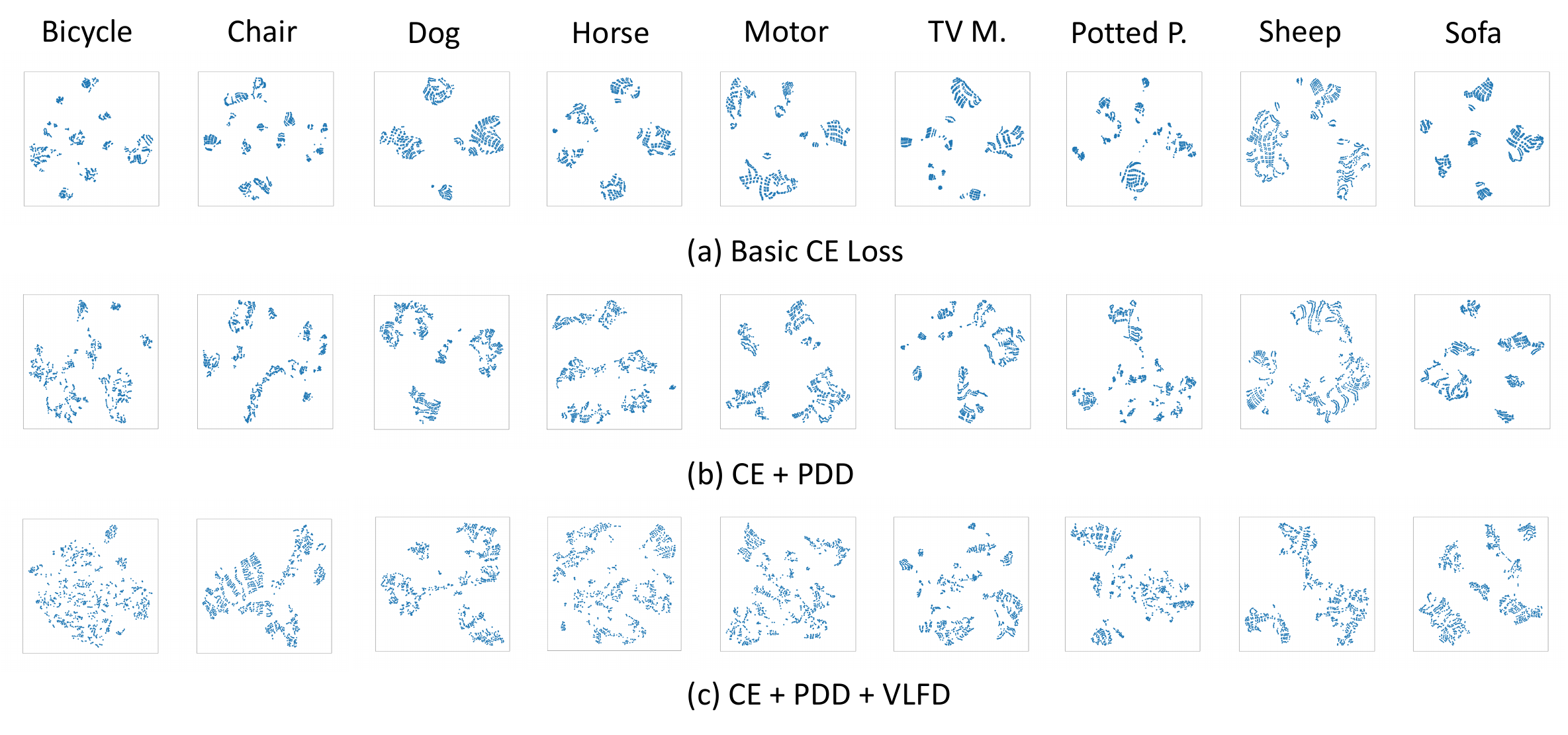}
    \caption{Per-class TSNE visualization results of adding our VLFD and PDD module on PASCAL VOC dataset.}
    \label{pdd_tsne}
\end{figure*}

Furthermore, we provide per-class TSNE visualizations in Fig.~\ref{pdd_tsne}. Our proposed PDD and VLFD methods encourage significantly better intra-class compactness in most of the categories, indicating that our proposed methods effectively capture and distill the discriminative features for each class. 

\begin{figure*}[t!]
    \centering
    \includegraphics[width=\textwidth]{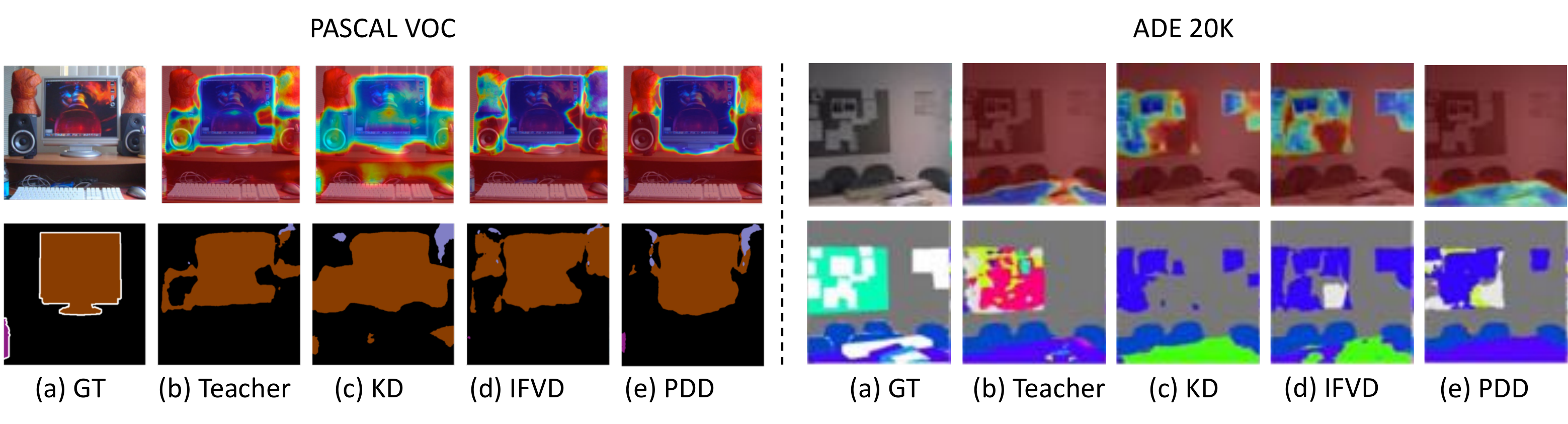}
    \caption{\textbf{Ablation study results for our PDD module}: (a) GT, (b) Teacher's prediction, (c) Vanilla KD, (d) IFVD, (e) PDD (ours).}
    \label{pdd_effect}
\end{figure*}

\subsection{The effectiveness of PDD module}

\begin{figure}[t!]
    \centering
    \includegraphics[width=\linewidth]{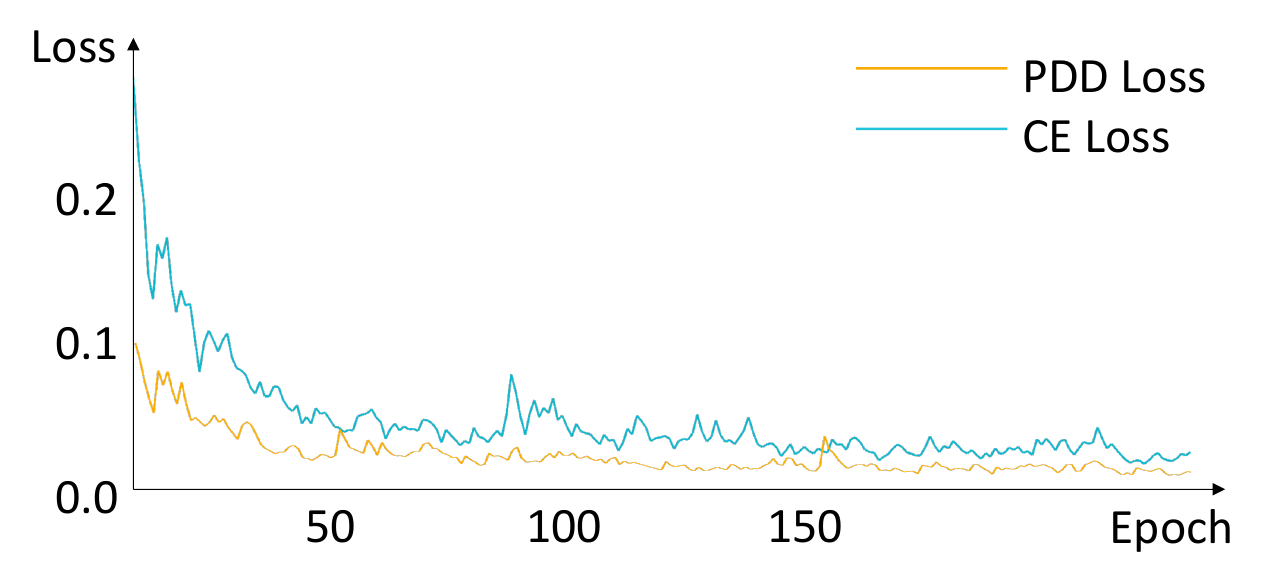}
    \caption{Loss curves of cross-entropy (CE) and our proposed $L_d$ loss.}
    \label{dkd_loss}
\end{figure}

We first provide the loss convergence curves of training with CE and our proposed PDD loss functions in Fig.~\ref{dkd_loss}. It is evident that our PDD loss $L_d$ achieves better loss convergence than the CE loss. These results demonstrate the effectiveness of our proposed PDD module in improving the performance of the student network.

Tab.~\ref{abloss} demonstrates that the student trained with the PDD loss $L_d$ outperforms the baseline (without KD) by \textbf{+0.69$\%$} in mIoU. Additionally, Fig.~\ref{pdd_effect} illustrates that our proposed PDD module effectively mitigates the errors and inadequacies of the teacher's predictions. This highlights the necessity of combining ground truth and teacher's predictions for accurate supervision. Notably, in Fig.~\ref{pdd_effect}, the student model trained with the PDD loss even outperforms the CNN teacher in terms of segmentation performance.

\begin{table}[t!]
\renewcommand{\tabcolsep}{3pt}
\resizebox{0.49\textwidth}{!}{
\begin{tabular}{cccccccc}
\toprule
Dataset                                                                & Loss                     & Baseline                  & \multicolumn{5}{c}{Distillation}                                                                                                          \\ \midrule
\multirow{6}{*}{\begin{tabular}[c]{@{}c@{}}VOC\end{tabular}} 
&$L_d$    & -        & $\checkmark$ & $\checkmark$ & $\checkmark$ & $\checkmark$ & $\checkmark$ \\ 
&$L_g$ & -        & -            & $\checkmark$ & -            & -            & $\checkmark$ \\ 
&$L_p$ & -        & -            & -            & $\checkmark$ & -            & $\checkmark$ \\ 
&$L_l$    & -        & -            & -            & -            & $\checkmark$ & $\checkmark$ \\ 
& \multicolumn{1}{l}{mIoU} & \multicolumn{1}{l}{69.51} & \multicolumn{1}{l}{70.20} & \multicolumn{1}{l}{70.53} & \multicolumn{1}{l}{70.35} & \multicolumn{1}{l}{70.42} & \multicolumn{1}{l}{70.76} \\ \cmidrule{2-8} 
& \multicolumn{1}{l}{$\Delta$}     & \multicolumn{1}{l}{base}  & \multicolumn{1}{l}{+0.69} & \multicolumn{1}{l}{+1.02} & \multicolumn{1}{l}{+0.84} & \multicolumn{1}{l}{+0.91} & \multicolumn{1}{l}{+1.25} \\ \midrule
Dataset                                                                & Loss                     & Baseline                  & \multicolumn{5}{c}{Distillation}                                                                                                          \\ \midrule
\multirow{6}{*}{CS}                                            
&$L_d$    & -        & $\checkmark$ & $\checkmark$ & $\checkmark$ & $\checkmark$ & $\checkmark$ \\ 
&$L_g$ & -        & -            & $\checkmark$ & -            & -            & $\checkmark$ \\ 
&$L_p$ & -        & -            & -            & $\checkmark$ & -            & $\checkmark$ \\ 
&$L_l$    & -        & -            & -            & -            & $\checkmark$ & $\checkmark$ \\ 
& \multicolumn{1}{l}{mIoU} & \multicolumn{1}{l}{72.52} & \multicolumn{1}{l}{72.87} & \multicolumn{1}{l}{73.10} & \multicolumn{1}{l}{73.17} & \multicolumn{1}{l}{73.50} & \multicolumn{1}{l}{73.58} \\ \cmidrule{2-8} 
                                                                       & \multicolumn{1}{l}{$\Delta$}     & \multicolumn{1}{l}{base}  & \multicolumn{1}{l}{+0.35} & \multicolumn{1}{l}{+0.58} & \multicolumn{1}{l}{+0.65} & \multicolumn{1}{l}{+0.98} & \multicolumn{1}{l}{+1.06} \\ \midrule
\end{tabular}}
\caption{\textbf{Ablation study of the distillation losses} on the PASCAL VOC 2012 (VOC) and Cityscapes (CS) datasets.}
\vspace{-4pt}
\label{abloss}
\end{table}

\subsection{Ablation of Loss Functions}
We study the effectiveness of the four losses $\mathcal{L}_d$, $\mathcal{L}_g$, $\mathcal{L}_p$, and $\mathcal{L}_l$ in our C2VKD framework. The baseline student model is trained with the basic Cross-Entropy (CE) loss without KD. In Tab.~\ref{abloss}, different combinations of losses are applied, and all the results are evaluated on the validation set of \textit{PASCAL VOC 2012} dataset. Meanwhile, we promote the supervised CE Loss to the PDD module for better pixel-wise image-level supervision.
As can be seen,  our PDD loss $L_d$, which transfers the knowledge based on the target and non-target class, achieves an improvement of mIoU by \textbf{+0.69$\%$} over the baseline. Moreover, $L_g$, $L_p$, and $L_l$ contribute positively to the mIoU with an increase of \textbf{+1.02$\%$}, \textbf{+0.84$\%$}, and \textbf{+0.91$\%$}, respectively.
We also study the effectiveness of the four losses of our C2VKD framework on the \textit{Cityscapes} dataset. In Tab.~\ref{abloss}, our PDD loss $L_d$, which transfers the disentangled pixel-wise knowledge, achieves an improvement of mIoU by +0.35$\%$ over the baseline. Meanwhile, $L_g$, $L_p$, and $L_l$ contribute positively to the mIoU with an improvement of \textbf{+0.58$\%$}, \textbf{+0.65$\%$} and \textbf{+ 0.98$\%$}, respectively.

\begin{table}[t!]
\centering
\setlength{\tabcolsep}{8pt}
\resizebox{\linewidth}{!}{
\begin{tabular}{ccccc}
\midrule
  & Base   & $\alpha$/$\beta$ = 3/1 & $\alpha$/$\beta$ = 2/1 & $\alpha$/$\beta$ = 1/1 \\ \midrule
mIoU  & 69.51 & 69.80     & 69.91     & \textbf{70.20}     \\ \midrule
$\Delta$ & -  & +0.39     & +0.40     & \textbf{+0.69}     \\ \midrule
\end{tabular}}
\caption{\textbf{Ablation study results of the $\alpha$ and $\beta$ in PDD} on the PASCAL VOC 2012 val set.}
\vspace{-4pt}
\label{alphabeta}
\end{table}

\subsection{Ablation of Hyper-parameters}
Tab.~\ref{alphabeta} reports the student model's mIoU$(\%)$ with different ratios of $\alpha$ and $\beta$ on the \textit{PASCAL VOC} val set. The baseline is trained using the CE loss, only focused on the knowledge on the target classes. As shown in Tab.~\ref{alphabeta}, when the importance $\beta$ for the non-target classes is increased, the student model learns more dark knowledge~\cite{hinton2015distilling} and achieves better performance. 

\subsection{Comparison of Computational Costs}
Note that no additional operations are added during inference, so the FLOPs and Params during inference are identical among all compared methods.
We also provide the FLOPs and Params comparison of all methods using the SegFormer-B0 model during training (without the teacher model's costs) in Tab.~\ref{tab:flops}. Though C2VKD introduces some computational cost (+0.02G in FLOPs and +0.57M in Params), it accelerates the KD from CNN to ViT while IFVD~\cite{wang2020intra} and CD~\cite{shu2021channel} even lead to worse KD in some cases.
\label{ablation}

\begin{table}[t!]
    \centering
    \setlength{\tabcolsep}{6pt}
    \small
    \resizebox{\linewidth}{!}{
    \begin{tabular}{cccc}
    \toprule
        Methods:  & Others & C2VKD (Ours) & $\Delta$ \\ \midrule
        FLOPs(G)   &  6.96 & 6.98 & +0.02 \\ 
        Params(M)  &  3.80  & 4.07 & +0.27 \\ 
        mIoU @ VOC w/ PVT   &  63.38 & 65.53 & \textbf{+2.01} \\ \midrule
    \end{tabular}}
     \caption{Comparison of computational costs during training.}
     \vspace{-4pt}
    \label{tab:flops}
\end{table}

\subsection{Comparison of Training Time}
The total iterations of all the compared methods and our C2VKD are the same, \ie, 40K on VOC, 50K on Cityscapes, and ADE20K. To compare the training time of different KD methods, we conduct experiments with ResNet-101 (teacher) and Segformer-B1 (student) on the ADE-20K dataset. As shown in Tab.~\ref{tab:time}, our C2VKD achieves the best segmentation performance (\textbf{40.91}) while only introducing \textbf{0.01}GPU days of training time, compared with CD~\cite{shu2021channel} (39.08).

\subsection{KD performance vs. varying teacher model size.}
We conduct ablation with different teacher models with SegFormer-B0 as the student model. The results are shown in Tab.~\ref{tab:teacher}. Apparently, the size of the teacher model seems not clearly influence the KD performance of our C2VKD, and it consistently outperforms the other KD methods.

\begin{table}[t!]
    \centering
    \setlength{\tabcolsep}{4.2pt}
    \resizebox{\linewidth}{!}{
    \begin{tabular}{ccccc}
    \toprule
        Methods:  & KD~\cite{hinton2015distilling} & IFVD~\cite{wang2020intra} & CD~\cite{shu2021channel} & C2VKD \\ \midrule
        mIoU  & 38.36 & 39.52 & 39.08 & \textbf{40.91} \\ 
        Time(GPU Days)  & 0.52 & 0.83 & 0.86 & 0.87 \\ \bottomrule
    \end{tabular}}
    \vspace{4pt}
    \caption{Comparison on training time with Seg-B0 on ADE20K.}
    \vspace{-4pt}
    \label{tab:time}
\end{table}

\begin{table}[t!]
    \centering
    \small
    \setlength{\tabcolsep}{8pt}
    \resizebox{\linewidth}{!}{
    \begin{tabular}{ccccc}
    \toprule
        Methods:  & KD & IFVD~\cite{wang2020intra} & CD~\cite{shu2021channel} & C2VKD \\ \midrule
        R-50  & 68.73 & 69.17 & 68.67 & \textbf{70.03} \\ 
        R-101  & 70.05 & 69.77 & 70.07 & \textbf{70.76} \\ \midrule
    \end{tabular}}
    \caption{Comparison on different-size teacher models.}
    \vspace{-4pt}
    \label{tab:teacher}
\end{table}

\section{Conclusion}
In this paper, we have proposed a novel KD framework, namely C2VKD, to learn a compact ViT-based (student) model from a pre-trained cumbersome yet high-performance CNN-based model (teacher). 
First, we propose the VLFD module that aligns visual and linguistic-compatible representations and achieves KD between the aligned features. Second, we propose the PDD module to allow the student to progressively learn more reliable knowledge from the teacher's predictions. Our C2VKD framework significantly outperforms the SoTA KD methods by a large margin.

\noindent \textbf{Limitation and future work:} 
We plan to improve the efficiency of feature transformations between the CNN-based teacher and ViT-based student. Another future direction is to explore how to reverse our C2VKD by distilling a compact CNN-based student from a cumbersome yet high-performance ViT-based teacher.

\clearpage
\bibliographystyle{IEEEtran}
\bibliography{main}
\vspace{-50pt}
\begin{IEEEbiography}[{\includegraphics[width=1in,height=1.2in,clip,keepaspectratio]{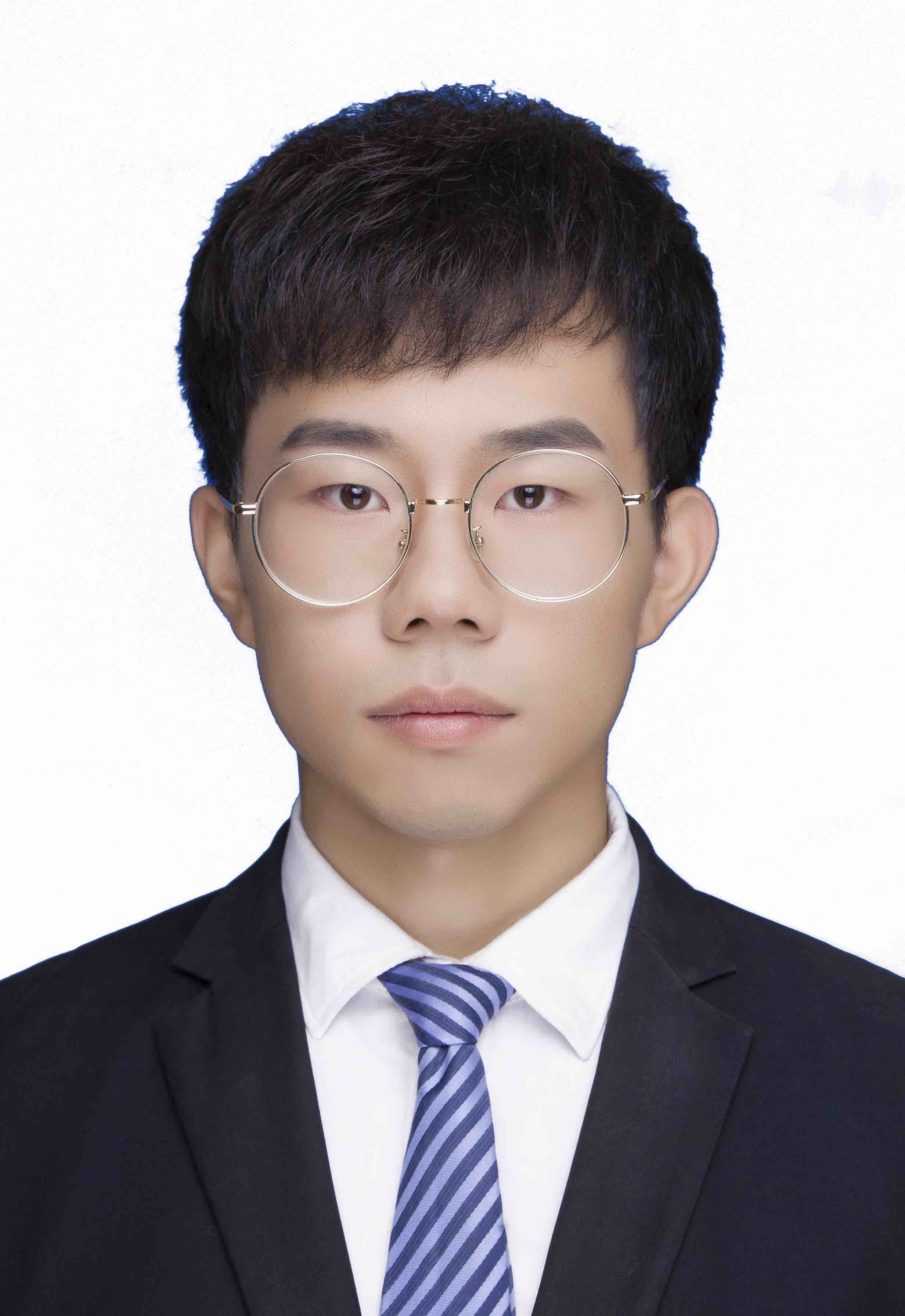}}] {Xu Zheng} (IEEE Student Member) is a Ph.D. student in the Visual Learning and Intelligent Systems Lab,  Artificial Intelligence
Thrust, The Hong Kong University of Science and Technology,  Guangzhou (HKUST-GZ). Before that, he obtained his B.E. and M.S. degree from Northeastern University, Shenyang, China.
His research interests lie in computer and robotic vision, multi-modal vision, vision language learning, \etc.
\vspace{-50pt}
\end{IEEEbiography}
\begin{IEEEbiography}[{\includegraphics[width=1in,height=1.2in,clip,keepaspectratio]{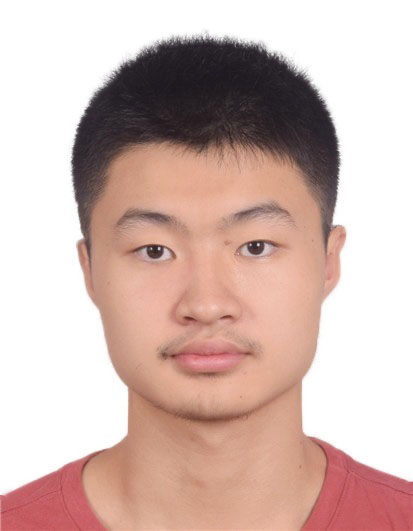}}] {Yunhao Luo}
is a M.S. student at Brown University. This work was done during his internship at the Visual Learning and Intelligent Systems Lab, HKUST-GZ. He obtained his B.E. degree from Southern University of Science and Technology (SUSTech). He has a broad research interest in Artificial Intelligence, including generative models, computer vision, decision-making.
\vspace{-50pt}
\end{IEEEbiography}

\begin{IEEEbiography}[{\includegraphics[width=1in,height=1.2in,clip,keepaspectratio]{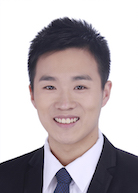}}] {Pengyuan Zhou}
 received his PhD from the University of Helsinki. He was a Europe Union Marie Curie ITN Early Stage Researcher from 2015 to 2018. He is currently a research associate professor at the School of Cyberspace Science and Technology, University of Science and Technology of
China (USTC). He is also a faculty member of the Data Space Lab, USTC. His research focuses on distributed networking AI systems, mixed reality
development, and vehicular networks.
\vspace{-50pt}
\end{IEEEbiography}

\begin{IEEEbiography}[{\includegraphics[width=1in,height=1.2in,clip,keepaspectratio]{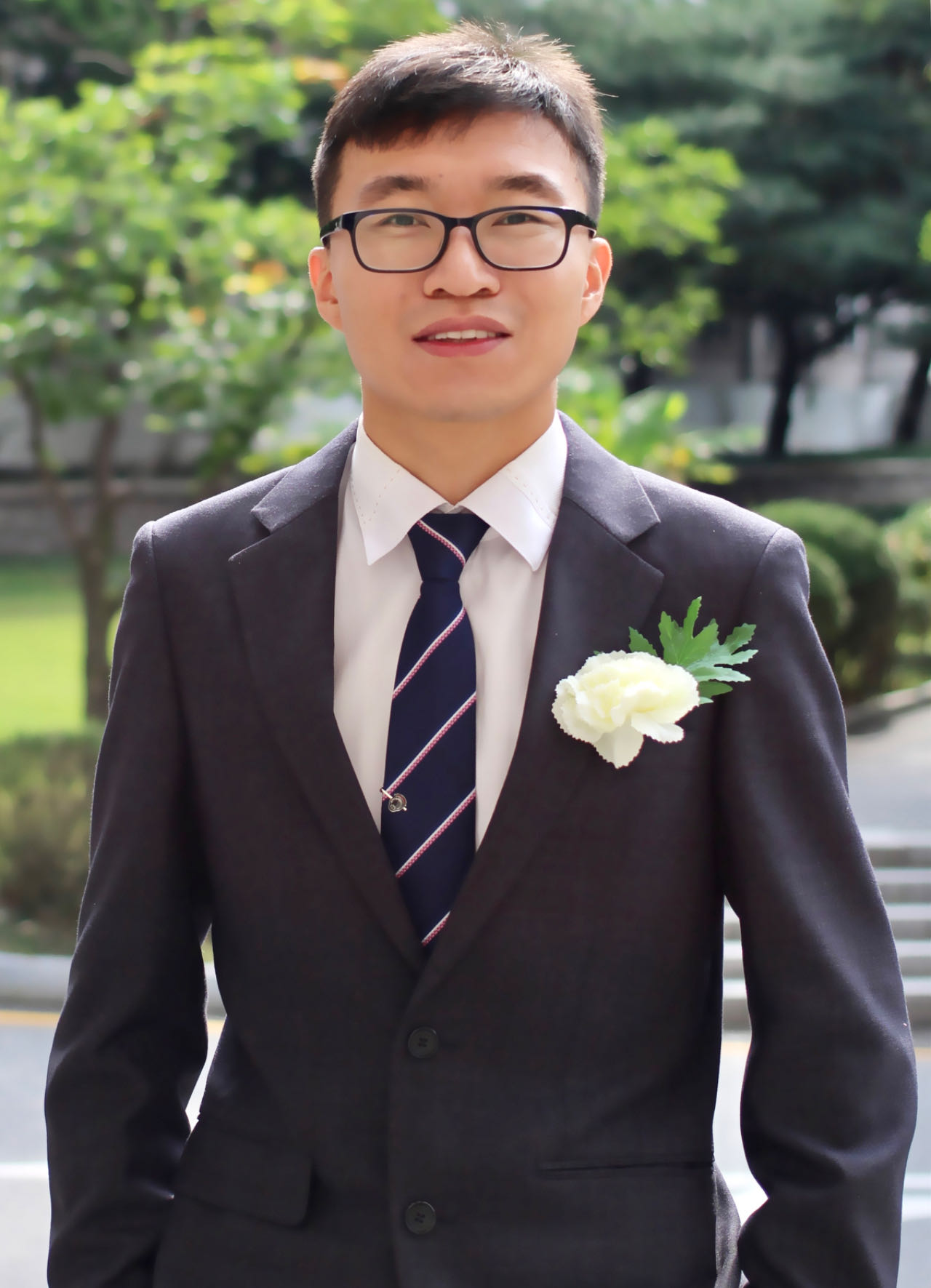}}] 
{Lin Wang} (IEEE Member) is an assistant professor in the AI Thrust and CMA Thrust, HKUST-GZ, and an affiliate assistant professor in the Dept. of CSE, HKUST. He is the director of the Visual Learning and Intelligent Systems(VLIS) Lab. He was a visiting researcher at the Imperial College London (ICL) from 2020-2021. He did his Postdoc at the Korea Advanced Institute of Science and Technology (KAIST). Before that, he got his Ph.D. (with honors) and M.S. from KAIST, Korea. He had rich cross-disciplinary research experience, covering mechanical, industrial, and computer engineering. His research interests lie in computer and robotic vision, machine learning, intelligent systems (XR, vision for HCI), etc. For more about me, please visit https://vlislab22.github.io/vlislab/.
\vspace{-50pt}

\end{IEEEbiography}

\end{document}